\documentclass[runningheads]{llncs}
\usepackage{graphicx}

\usepackage[style=numeric]{biblatex}
\usepackage{xcolor}

\usepackage[colorlinks=true, linkcolor=black, citecolor=black, urlcolor=black]{hyperref}

\usepackage{booktabs} 
\usepackage{adjustbox} 
\usepackage{multirow} 

\usepackage{float}
\usepackage{subcaption}
\usepackage{amsfonts}
\usepackage{amsmath}
\usepackage{lineno}
\usepackage{afterpage}
\usepackage{orcidlink}

\begin{document}

\title{DELINE8K: A Synthetic Data Pipeline for the Semantic Segmentation of Historical Documents}

\author{Taylor Archibald\inst{1}\orcidlink{0000-0003-2576-208X} \and
Tony Martinez\inst{1}}

\titlerunning{DELINE8K}
\authorrunning{Archibald and Martinez}

\institute{Brigham Young University, Provo, UT\\
\email{tarch@byu.edu, martinez@cs.byu.edu}\\
\url{https://axon.cs.byu.edu}
}
\maketitle
\begin{abstract}

Document semantic segmentation is a promising avenue that can facilitate document analysis tasks, including optical character recognition (OCR), form classification, and document editing. Although several synthetic datasets have been developed to distinguish handwriting from printed text, they fall short in class variety and document diversity. We demonstrate the limitations of training on existing datasets when solving the National Archives Form Semantic Segmentation dataset (NAFSS), a dataset which we introduce. To address these limitations, we propose the most comprehensive document semantic segmentation synthesis pipeline to date, incorporating preprinted text, handwriting, and document backgrounds from over 10 sources to create the \textbf{D}ocument \textbf{E}lement \textbf{L}ayer \textbf{IN}tegration \textbf{E}nsemble 8K, or DELINE8K dataset\footnote{DELINE8K is available at \href{https://github.com/Tahlor/deline8k}{https://github.com/Tahlor/DELINE8K}}. Our customized dataset exhibits superior performance on the NAFSS benchmark, demonstrating it as a promising tool in further research.


\end{abstract}


\section{Introduction}
\begin{figure}[h]
  \centering
  \subfloat[The handwriting has been tinted by class (handwriting=red, preprinted text=green, and grid lines=blue).]{\includegraphics[width=.47\textwidth]{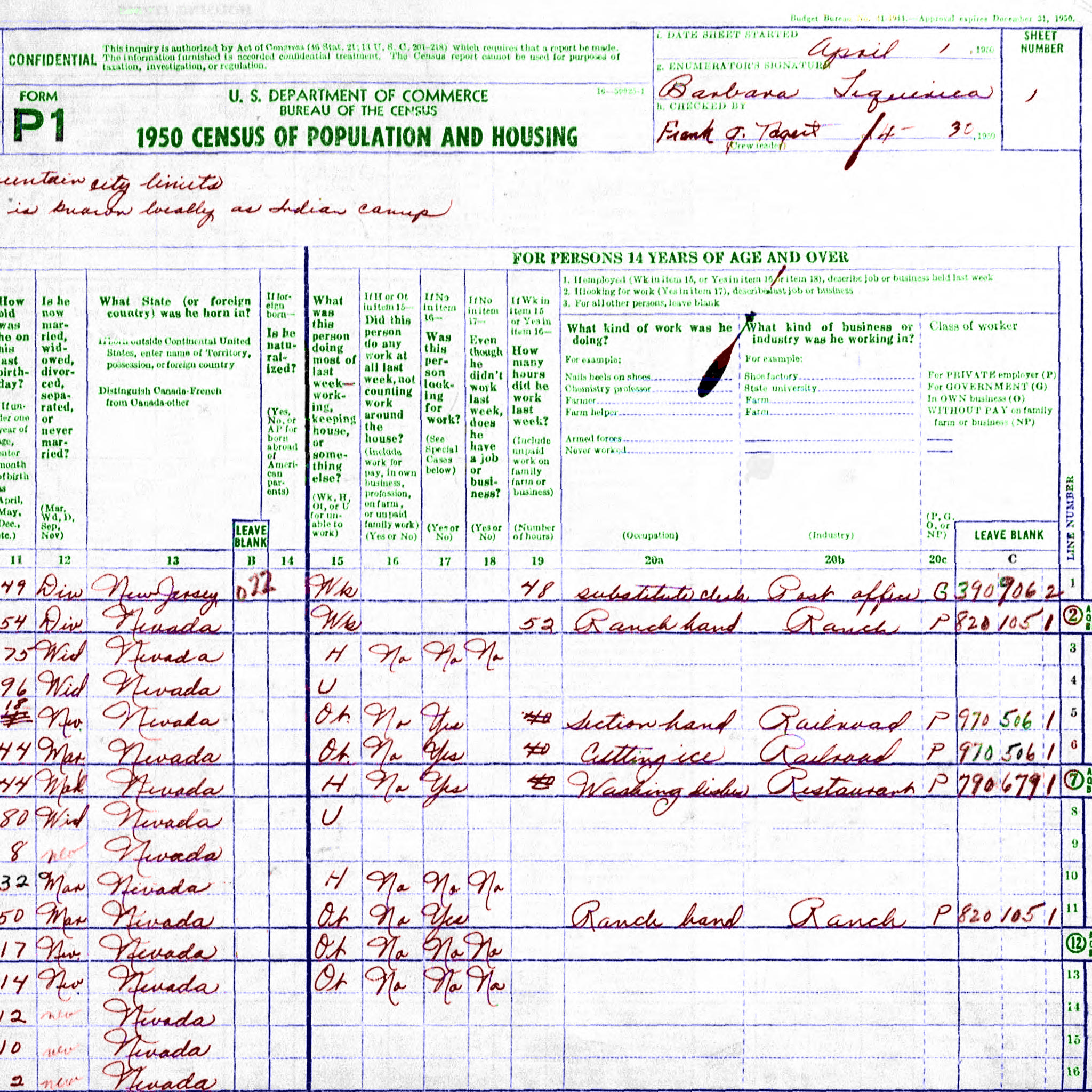}}
  \hspace{5mm}
  \subfloat[The handwritten elements have been removed and replaced using content aware in-painting. Detected grid lines were overlaid to recover faded lines.]{\includegraphics[width=0.47\textwidth]{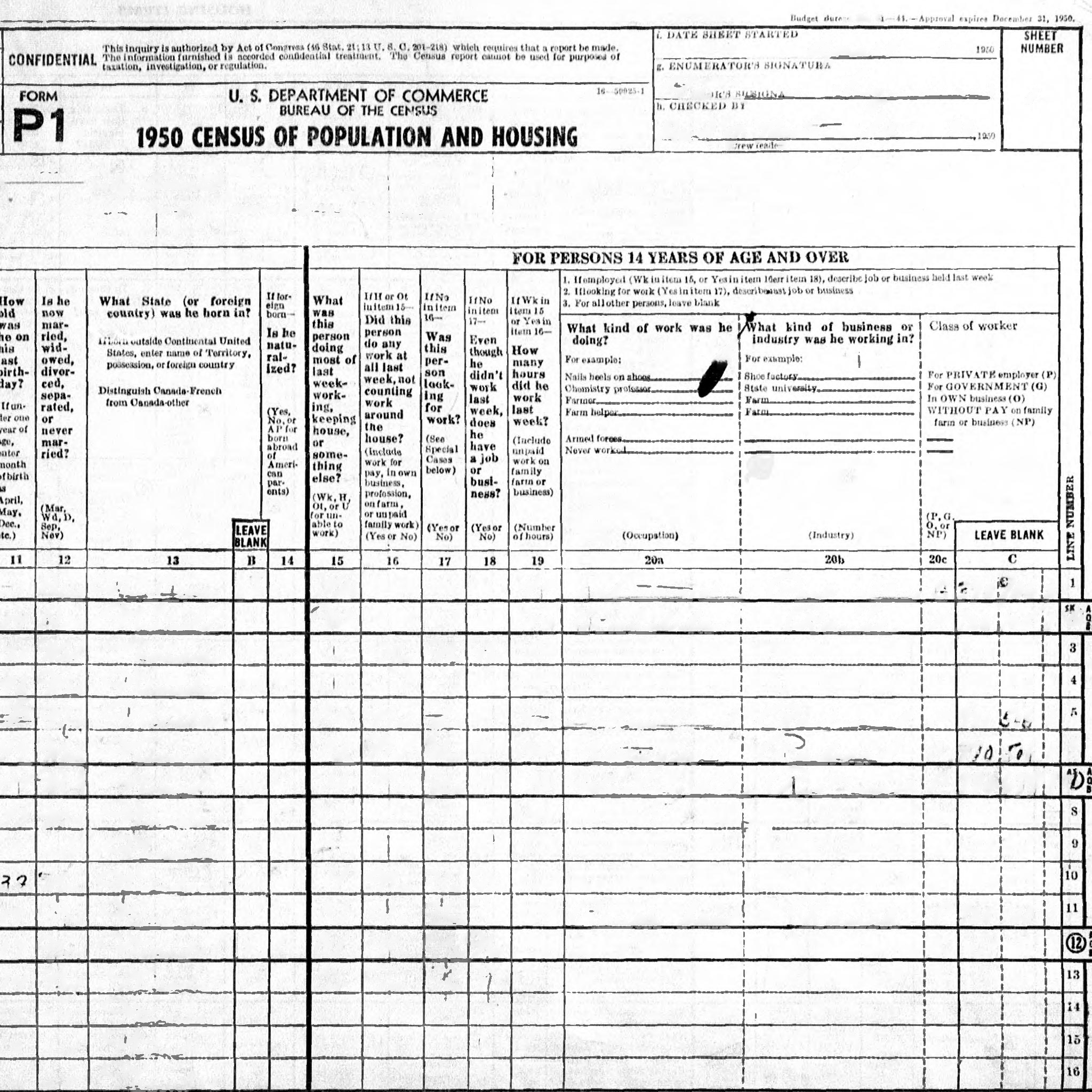}}
  \caption{A 1950's US Census Form can be decomposed into different content classes using a model trained on DELINE8K.}
  \label{fig:blanked}

\end{figure}

Once ubiquitous office supplies like whiteout are anachronisms in the digital age. Although the use of physical whiteout has diminished, we sometimes need to perform similar corrections digitally while preserving document content. Perhaps one has a completed form that requires redaction or needs to be completely stripped of handwritten content, as illustrated in Figure~\ref{fig:blanked}. Or, for data scientists working with a document collection, they may require a pristine document for a template matching task, want to eliminate extraneous markings to improve optical character recognition (OCR), or identify blank forms to exclude from a collection.

One avenue to realize these ambitions is through semantic segmentation. Semantic segmentation involves partitioning images into distinct layers corresponding to distinct objects. A specific, extensively studied form of document semantic segmentation, known as binarization, involves distinguishing and separating the textual content from the background, effectively converting the document into a binary image of black and white pixels. This technique is pivotal for enhancing the clarity and readability of documents, thereby facilitating tasks such as template matching and OCR enhancement by removing irrelevant markings or background noise. 

The binarization task can be extended to a multiclass segmentation problem, which can entail differentiating among various document components including handwritten notes, printed text, form elements, stamps, images, and signatures. Previous efforts~\cite{joHandwrittenTextSegmentation2020,vafaieHandwrittenPrintedText2022, gholamianHandwrittenPrintedText2023} have focused on distinguishing between preprinted elements and handwritten ones through the creation of synthetic datasets by superimposing handwritten and preprinted document images.

While this approach shows promise, there are potential pitfalls in producing a single universal dataset. For instance, the definition of what distinguishes handwritten from printed text may depend on the context. For some tasks, it may be necessary to identify elements that have been stamped or typewritten. Some collections may have unique fonts not captured in a standardized training dataset. Factors such as text rotation and the degree of document degradation further complicate the development of universally applicable models. While new training data can be labeled for each new collection, this aggregate expense would be considerable. These considerations underscore the necessity for a nuanced approach to dataset creation, one that accommodates the diverse and specific needs of the dataset being segmented.

We posit that our approach to generating synthetic data is more comprehensive and better suited to the diverse and complex nature of documents encountered in many real-world scenarios. In light of the myriad document types—ranging from historical manuscripts to modern fillable forms, each with its unique font styles, elements, and document backgrounds—it becomes clear that a one-size-fits-all solution is impractical. Consequently, we argue that it is more advantageous to develop custom synthetic data that closely mirrors the specific characteristics of the documents requiring segmentation.

To address this need, we introduce a suite of synthetic tools~\cite{docgen} designed to facilitate the creation of tailored semantic segmentation document datasets. This suite includes curated document backgrounds, form elements, and a diverse array of text and non-text components, which enable users to simulate a wide range of document conditions and complexities. 

We identify the National Archives Forms dataset (NAF)~\cite{davisDeepVisualTemplateFree2019} as presenting an unsolved challenge in the realm of document semantic segmentation. This dataset is particularly complex due to its inclusion of historical documents, which exhibit significant variations in scanning quality and document condition. Because it consists of many documents with preprinted form elements, including blank tables and baselines below handwritten fields, we propose classifying these form elements as separate from either preprinted text or handwriting, which may prove useful in downstream tasks. To benchmark performance, we introduce the National Archives Forms Semantic Segmentation dataset (NAFSS) test set, with samples drawn from the NAF dataset. To solve NAFSS, we create the DELINE8K dataset with our synthetic pipeline, composed of 8,000 $768\times768$ images. While we argue the DELINE8K dataset is, in many ways, more comprehensive than previous synthetic semantic segmentation datasets, it primarily serves as a template for the creation of synthetic data tailored to new document collections.

\section{Related Work}
Binarization, a well-studied problem that goes back decades, is a special case of semantic segmentation, and aims to classify pixels into two categories: foreground or background, producing a binary representation of an image. The Document Image Binarization Contest (DIBCO) series, running most years from 2009 to 2019, has been instrumental in pushing forward the development and evaluation of document binarization techniques. This contest provides researchers with a platform to benchmark their algorithms against a diverse set of challenging document images, including historical manuscripts and printed texts with varying degrees of degradation. The progression of DIBCO over the years has not only showcased the evolution of binarization methods but also highlighted the increasing effectiveness of deep learning approaches in this domain. Deep learning-based methods, particularly convolutional neural networks 
(CNNs)~\cite{tensmeyerDocumentImageBinarization2017,ronnebergerUNetConvolutionalNetworks2015}, have demonstrated strong performance in handling complex variations in text and background, adapting to different scripts, and managing noise and artifacts better than traditional thresholding and edge detection techniques. An example of this was demonstrated in \cite{ronnebergerUNetConvolutionalNetworks2015}, which employed a CNN-based architecture from medical imaging known as the U-Net~\cite{ronnebergerUNetConvolutionalNetworks2015} and won the 2017 Document Image Binarization (DIBCO) competition~\cite{pratikakisICDAR2017CompetitionDocument2017}.

Multiclass semantic segmentation represents a natural extension of binarization, expanding the binary classification problem to encompass multiple categories, thereby enabling the detailed partitioning of document images into varied semantic regions beyond the simple foreground and background classes. These categories may include printed text, handwritten text, images, stamps, form elements, etc. The feasibility of this task was demonstrated in~\cite{stewartDocumentImagePage2017}, though the study was very narrow in scope, using only a single training image. The cumbersome nature of labeling documents for semantic segmentation, along with a dearth of publicly accessible datasets, represents a significant impediment to this line of research.

This challenge has been addressed primarily through generating synthetic datasets by superimposing handwritten and preprinted documents~\cite{joHandwrittenTextSegmentation2020,vafaieHandwrittenPrintedText2022, gholamianHandwrittenPrintedText2023}. \cite{vafaieHandwrittenPrintedText2022}~introduces the WGM-SYN dataset, with printed text taken from ``Pilotprojekt zur Wiedergutmachung'' archival documents, 
and \cite{gholamianHandwrittenPrintedText2023} introduces SignaTR6K, which contains crops of legal documents with superimposed signatures. 

\begin{figure}[H]
  \centering
  {\includegraphics[width=.55\textwidth]{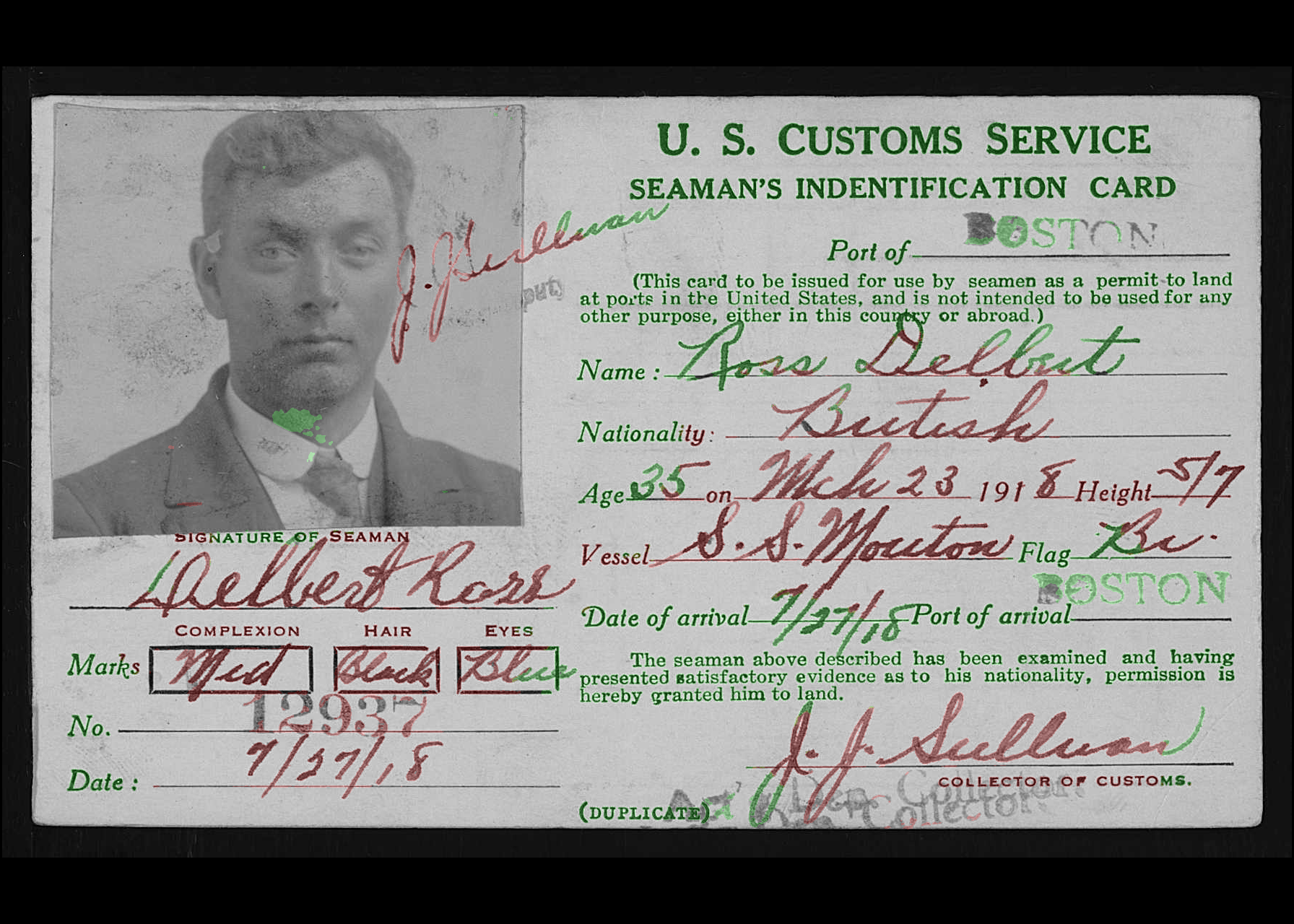}}
  \hspace{5mm}
  \caption{A model trained on DIBCO exhibits pockets of success on the task of distinguishing text from handwriting. Insufficient examples where both text and handwriting are present in the image may contribute to the missed classifications. }
  \label{fig:DIBCO_FAILURE}

\end{figure}

Notably, these datasets do not incorporate the wealth of varied backgrounds and degradations that some recent binarization datasets do, including \cite{groleauShabbyPagesReproducibleDocument2023} and \cite{sadekarLSHDIBLargeScale2022}. Figure~\ref{fig:DIBCO_FAILURE} shows unsatisfactory results when training a segmentation model that distinguishes between text and handwriting on DIBCO, likely because it lacks instances that feature both classes in the same image. Moreover, all of these datasets only have labels for two classes, handwriting and preprinted text.

We propose greatly expanding the synthesis pipeline to extend these approaches to work more effectively on historical documents, in addition to expanding the number of classes.

\section{Method}
\subsection{Problem Formalization}

Given a document image $I \in \mathbb{R}^{H \times W \times 3}$, where $H$ and $W$ are the height and width of the image respectively, our goal is to segment this image into $C$ different classes. These classes can represent semantic components such as handwritten text, printed text, or form elements. 

The task is to learn a mapping function $F: \mathbb{R}^{H \times W \times 3} \rightarrow \mathbb{R}^{H \times W \times C}$, which assigns each pixel in $I$ to one of $C$ classes. The output of $F(I)$ is a set of $C$ probability maps indicating the likelihood of each pixel belonging to each class. Each probability is rounded to either 0 or 1 before computing each evaluation metric.

To train a model that approximates $F$, we use a synthetic dataset\nopagebreak
\begin{equation}
\mathcal{D}_{\text{synth}} = \{(I'_i, L'_i)\}_{i=1}^{N'}, 
\end{equation}\nopagebreak
where each $I'_i$ is a synthetic input image, and $L'_i \in \{1,0\}^{H \times W \times C}$ is the corresponding synthetic ground truth, where $\{1,0\}$ represent positive and negative class instances, respectively. This dataset $\mathcal{D}_{\text{synth}}$ is designed to mimic the distribution of some real dataset $\mathcal{D}_{\text{real}}$, allowing us to train the model in environments where collecting large amounts of real annotated data may be impractical. For each class of interest, including handwriting, printed text, and form elements, in addition to the non-class document background, we acquire a dataset that contains the element of interest, i.e., $\{\mathcal{D}_{\text{hw}},\mathcal{D}_{\text{text}},\mathcal{D}_{\text{form}}, \mathcal{D}_{\text{bg}}\}$.

To construct $\mathcal{D}_{\text{synth}}$, we begin with an image from one of the component classes, e.g., $I'_{bg} \in \mathcal{D}_{\text{hw}}$, which contains a document background and none of the other classes. We then consider an image from of the other classes, e.g., $I'_{hw}$, and assign a ground truth label $L'_{hw_{ij}} \in \{0, 1\}$ for each pixel based on a threshold $\tau$ for pixel intensity to determine sufficient darkness. The composite image creation involves overlaying $I'_{hw}$ onto a background image $I'_{bg}$, derived from a separate dataset. The compositing process is mathematically represented as follows:\nopagebreak


\[
C_{hw} = \left( \frac{I'_{hw}}{\max(I'_{hw})} \right) \cdot I'_{bg} \quad,
\]
\nopagebreak
where $C_{hw}$ is the resulting composite image, and $\max(I'_{hw})$ normalizes the handwriting intensity to the range $[0, 1]$ before multiplication with the background $I'_{bg}$ to ensure that each class appears darker than its background. We found this approach to yield more consistent results than seamless cloning~\cite{perezPoissonImageEditing2003}. 


To train our model, we aim to minimize the Dice loss, defined as:\nopagebreak

\begin{equation}
\mathcal{L}_{\text{Dice}} = 1 - \frac{1}{C} \sum_{c=1}^C \text{Dice}_c.
\end{equation}
\\
\newpage
The Dice coefficient is defined for class $c$ as:\nopagebreak
\begin{equation}
\text{Dice}_c = \frac{2 \times \sum_{i,j} p_{c_{ij}} \cdot y_{c_{ij}}}{\sum_{i,j} p_{c_{ij}}^2 + \sum_{i,j} y_{c_{ij}}^2},
\end{equation}\nopagebreak
where $p_{c_{ij}}$ is the predicted probability of the pixel at position $(i, j)$ belonging to class $c$, and $y_{c_{ij}}$ is a binary indicator (0 or 1) if class label $c$ is the correct classification for the pixel at position $(i, j)$. This formulation essentially measures the overlap between the predicted probabilities and the ground truth labels, acting as a harmonic mean of precision and recall for each class. Consequently, this loss function is particularly effective for document semantic segmentation as it controls for class imbalances.

\subsection{Datasets}

\subsubsection{DELINE8K}

A preeminent challenge of semantic segmentation is that ground truth labels are often contingent upon the broader context of the problem at hand. Some context-dependent challenges include:
\begin{itemize}
\item distinguishing between printed, typewritten, and stamped text, when they may use the same fonts;
\item distinguishing between handwriting and fonts designed to mimic handwriting; 
\item segmenting baselines, or the preprinted reference lines beneath handwritten text in a document, which are often excluded from the foreground ground truth (e.g., see the 2012 DIBCO competition images~\cite{pratikakisICFHR2012Competition2012});
\item segmenting text within logos, where characters might be artistically altered or merged with graphic elements.
\end{itemize}

While solving these cases above may be feasible within a uniform, narrowly defined dataset, optimizing for them when it is not required may impair generalization across a broader range of datasets.

To address these variations, we propose synthesizing datasets that are aligned with the characteristics of a target dataset and task requirements. This leads us to create the DELINE8K dataset, a dataset comprised of 8,000 $768\times768$ images with up to four layers: background, handwriting, printed text, and form elements, created specifically to solve the NAF and similar datasets. Below, we detail the composition of each layer.

\paragraph{Backgrounds}

For the background layer, we synthesized over 10,000 images using DALL$\cdot$E~\cite{rameshZeroShotTexttoImageGeneration} to create a wide variety of textures and document backgrounds that mimic real-world documents, as depicted in Figure~\ref{tab:example_images}. We intentionally create some images where a document is in the foreground set against some non-document background. We use Doc-UFCN from~\cite{boilletMultipleDocumentDatasets2021} to identify valid document regions on to which to composite other layers.

\begin{table}[H]
\centering
\begin{tabular}{cc}
  \includegraphics[width=0.375\textwidth]{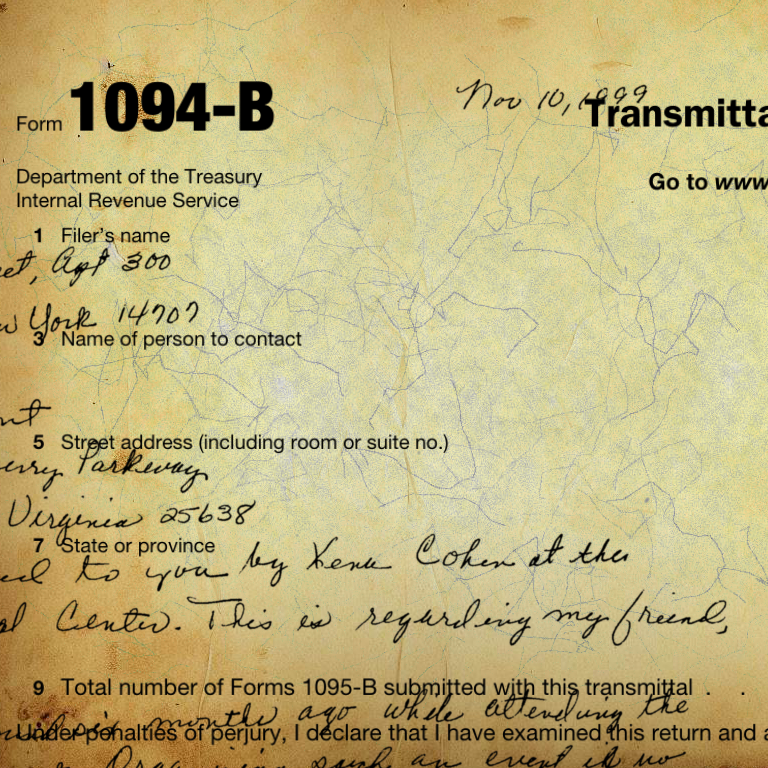} & 
  \includegraphics[width=0.375\textwidth]{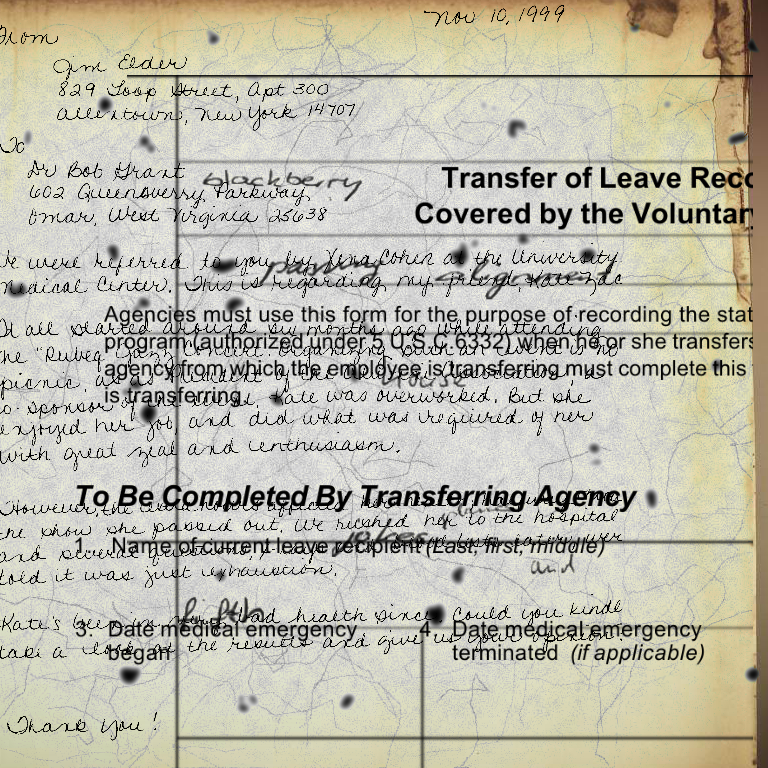} \\
  (a) & (b) \\[6pt]
  \includegraphics[width=0.375\textwidth]{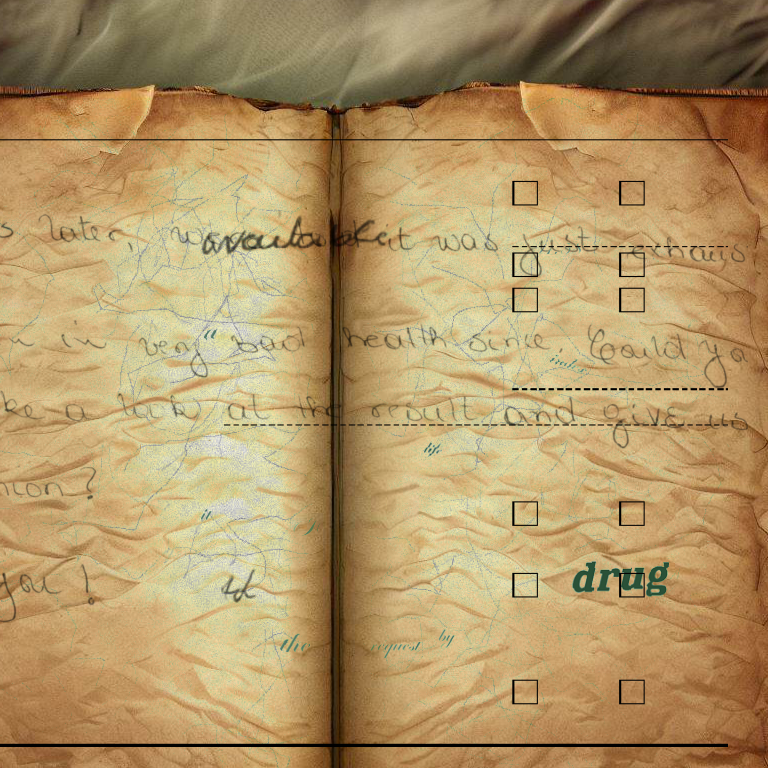} & 
  \includegraphics[width=0.375\textwidth]{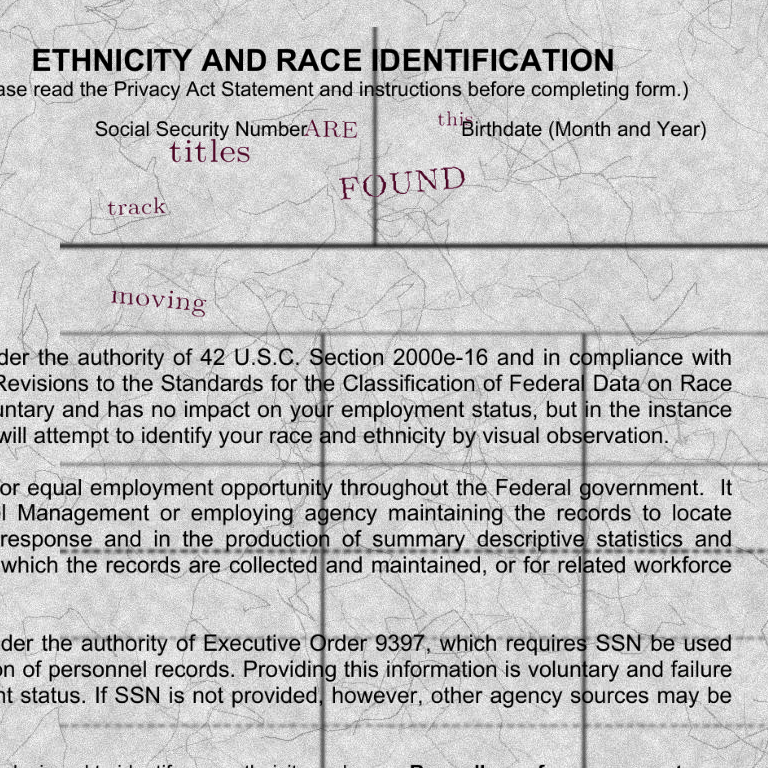} \\
  (c) & (d) \\
\end{tabular}
\caption{Example DELINE8K images with DALL$\cdot$E backgrounds.}
\label{tab:example_images}
\end{table}

\paragraph{Handwriting}

The handwriting component amalgamates data from several sources, including the IAM database~\cite{martiIAMdatabaseEnglishSentence2002}, CSAFE Handwriting Database~\cite{crawfordDatabaseHandwritingSamples2020}, EMNIST~\cite{cohenEMNISTExtendingMNIST2017}, and CEDAR-LETTER ~\cite{hullDatabaseHandwrittenText1994}. Additionally, to augment the variability and volume of handwriting examples, we use Handwriting Transformers, a handwriting synthesis model proposed in~\cite{bhuniaHandwritingTransformers2021}, that was trained on the IAM~\cite{martiIAMdatabaseEnglishSentence2002} and the CVL Database~\cite{kleberCVLDataBaseOffLineDatabase2013}, to generate unique handwriting samples that extend the dataset's coverage.

\paragraph{Printed Text}

Printed text samples were curated from a broad spectrum of fonts, with over 10,000 fonts sourced from \href{http://www.1001freefonts.com}{1001freefonts.com}~\cite{freeFonts}. We also extract text data from over 10,000 forms obtained from U.S. government agencies using PyPDF. These forms were scraped from the IRS~\cite{irsForms}, OPM~\cite{opmForms}, SSA~\cite{ssaForms}, and GSA~\cite{gsaForms}, which collectively provide a rich foundation for simulating real-world document structures and layouts.

\paragraph{Form Elements}

We similarly extract form elements from the U.S. government agency forms above~\cite{irsForms,opmForms,ssaForms,gsaForms}. Because these documents often have large blocks of colored background, we exclude any document with a $10\times10$ pixel region that is entirely non-white, as our present  interest is in capturing grid lines, check boxes, and printed baselines. We generate additional grid patterns using matplotlib.

\subsubsection{DIBCO}

We train a model on all DIBCO competition datasets (with the exception of the special 2019 handwriting split that contains ancient documents). The ground truth for each image is considered to be either entirely handwritten or preprinted. The only exception is image \#4 of the 2019 competition, where we label the stamped portion as printed text and use the handwriting label elsewhere.

\subsubsection{SignaTR6K}

We train a model on the SignaTR6K dataset~\cite{gholamianHandwrittenPrintedText2023}, which consists of $256\times256$ pixel crops. The dataset is divided into a training set with 5,169 images, a validation set containing 530 images, and a test set comprising 558 images.

\subsubsection{NAF}

We selected 3 images randomly from the NAF dataset~\cite{davisDeepVisualTemplateFree2019} dataset to label. For convenience, these were padded so each dimension is divisible by 256, giving the dataset a total of 503 $256\times256$ mutually exclusive patches. All metrics are reported using the simple average of the scores computed for each of the original 3 images after padding.

\subsubsection{Data Augmentations}

To simulate a wide range of real-world document and image artifacts, we use Albumentations~\cite{buslaevAlbumentationsFastFlexible2020}, Augraphy~\cite{theaugraphyprojectAugraphyAugmentationPipeline2023}, and the NVlabs ocrodeg library~\cite{ocrodeg} to perform rotation, scaling, noising, and blurring. A detailed review of all augmentations used can be found in the supplemental materials, Section~\ref{augmentation} (\nameref{augmentation}).

\subsection{Architecture}

For all of our experiments, we use a U-Net~\cite{ronnebergerUNetConvolutionalNetworks2015} with a ResNet50~\cite{heDeepResidualLearning2015} encoder, with a combined 32.5M parameters.

\section{Results}

\subsection{Metrics}
In assessing the performance of our document semantic segmentation approach, we utilize several metrics to capture the accuracy and reliability of our method. The F-measure, the harmonic mean of precision and recall, provides a balanced view of the model's effectiveness in identifying relevant text segments. We also report the pseudo-F-measure, where recall is calculated using a skeletonized ground truth~\cite{ObjectiveEvaluationMethodologyBinarization}, a metric intended to evaluate the model based on its ability to accurately capture the essential structural details of the text while minimizing the impact of variations in text thickness and style. Another common semantic segmentation metric we report is Intersection over Union (IoU), which is a geometric comparison between the predicted and actual class regions. Lastly, we report the percent of false positives for images without positive ground truth values, since the other metrics in these cases are either always 0 or undefined (e.g., consider the case when a model detects preprinted text on a blank page). This tailored approach ensures a nuanced assessment of model performance across diverse document types and conditions.

\begin{figure}[h]
\centering

\begin{tabular}{ccccc}
\includegraphics[width=0.18\textwidth]{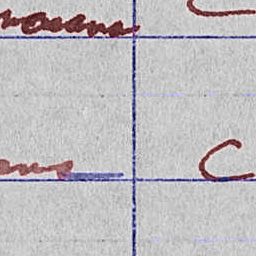} &
\includegraphics[width=0.18\textwidth]{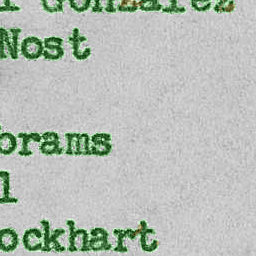} &
\includegraphics[width=0.18\textwidth]{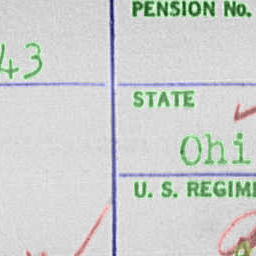} &
\includegraphics[width=0.18\textwidth]{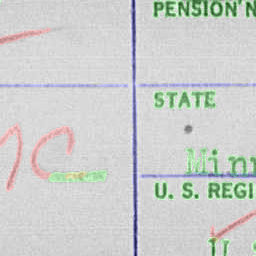} &
\includegraphics[width=0.18\textwidth]{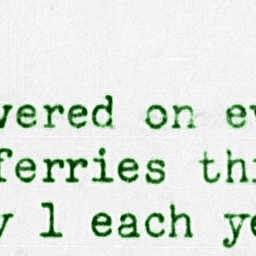} \\
\includegraphics[width=0.18\textwidth]{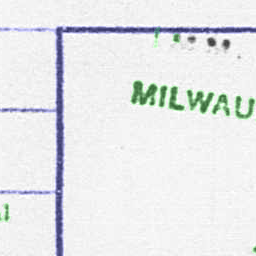} &
\includegraphics[width=0.18\textwidth]{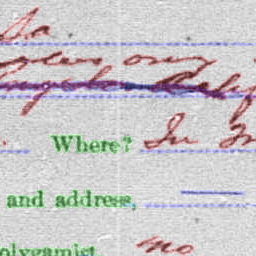} &
\includegraphics[width=0.18\textwidth]{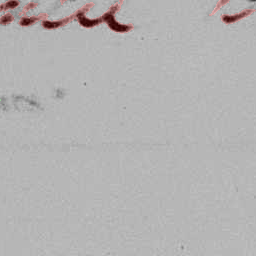} &
\includegraphics[width=0.18\textwidth]{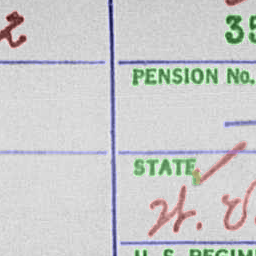} &
\includegraphics[width=0.18\textwidth]{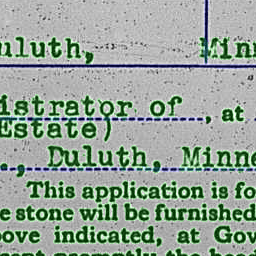} \\
\includegraphics[width=0.18\textwidth]{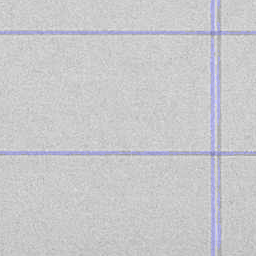} &
\includegraphics[width=0.18\textwidth]{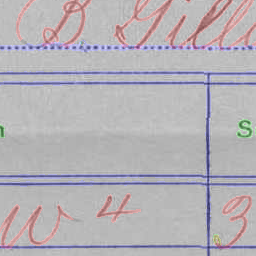} &
\includegraphics[width=0.18\textwidth]{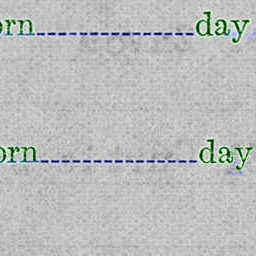} &
\includegraphics[width=0.18\textwidth]{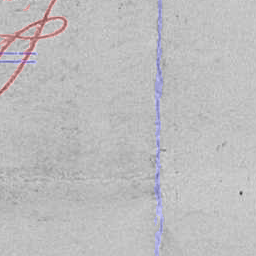} &
\includegraphics[width=0.18\textwidth]{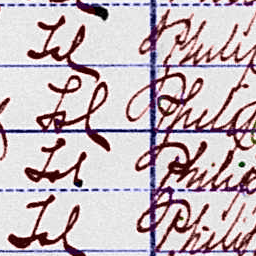} \\
\includegraphics[width=0.18\textwidth]{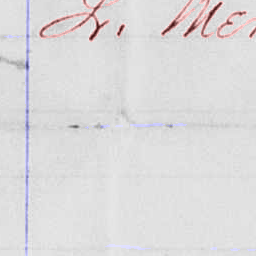} &
\includegraphics[width=0.18\textwidth]{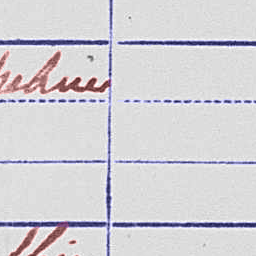} &
\includegraphics[width=0.18\textwidth]{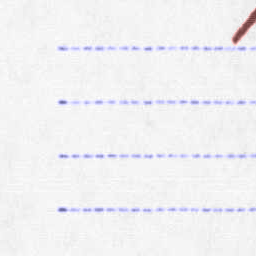} &
\includegraphics[width=0.18\textwidth]{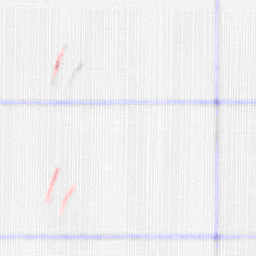} &
\includegraphics[width=0.18\textwidth]{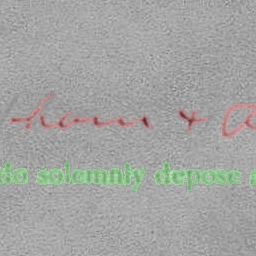} \\
\includegraphics[width=0.18\textwidth]{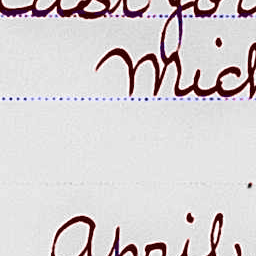} &
\includegraphics[width=0.18\textwidth]{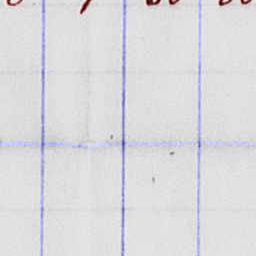} &
\includegraphics[width=0.18\textwidth]{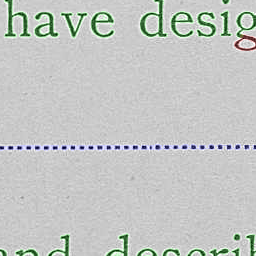} &
\includegraphics[width=0.18\textwidth]{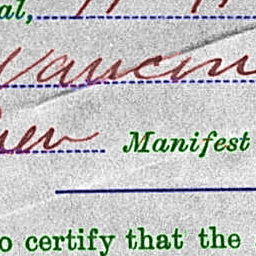} &
\includegraphics[width=0.18\textwidth]{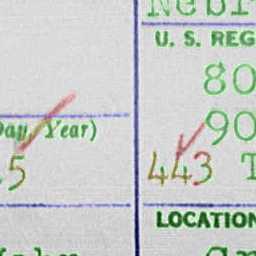} \\
\end{tabular}
\caption{Evaluation on randomly selected patches of the NAF dataset, excluding blanks}
\label{fig:NAF_results}
\end{figure}

\subsection{Experiments}
We train a UNet on SignaTR6K, DELINE8K, and the DIBCO competition datasets, and evaluate on the SignaTR6K test split, our NAFSS dataset, and the DIBCO datasets. We report segmentation metrics for handwriting and printed (TEXT) classes for each training dataset and test dataset combination in Table~\ref{tab:MAIN_TABLE}. Because only NAFSS and DELINE8K include a form element class (FORM), we also report evaluations of the SignaTR6K and DIBCO models comparing their preprinted text predictions against the combined preprinted text and form element classes (TEXT+FORM).

\begin{table}[!h]
\centering
\caption{Performance across training and evaluation datasets}
\label{tab:MAIN_TABLE}
\begin{adjustbox}{center, max width=\textwidth}
\begin{tabular}{l@{\hspace{5mm}}l@{\hspace{5mm}}l@{\hspace{5mm}}rr@{\hspace{5mm}}rr}
\toprule
Evaluation Dataset & Class & Training Dataset & F-measure & pF-measure & IoU & \% False Positive \\
\midrule
\multirow{4}{*}{DIBCO}
 & \multirow{2}{*}{HANDWRITING} & SignaTR6K & \textbf{0.793} & \textbf{0.849} & \textbf{0.688} & 1.5 \\
 &  & DELINE8K & 0.698 & 0.7 & 0.565 & 4.2 \\
\cmidrule{2-7}
 & \multirow{2}{*}{TEXT} & SignaTR6K & \textbf{0.718} & \textbf{0.773} & \textbf{0.578} & 1.9 \\
 &  & DELINE8K & 0.682 & 0.692 & 0.534 & 0.6 \\
\cmidrule{1-7}
\multirow{10}{*}{NAFSS}
 & \multirow{3}{*}{HANDWRITING} & DELINE8K & \textbf{0.876} & 0.877 & \textbf{0.785} &  \\
 &  & SignaTR6K & 0.821 & \textbf{0.885} & 0.698 &  \\
 &  & DIBCO & 0.77 & 0.813 & 0.627 &  \\

 \cmidrule{2-7}
 & \multirow{3}{*}{TEXT} & DELINE8K & \textbf{0.917} & \textbf{0.914} & \textbf{0.847} &  \\
 &  & DIBCO & 0.756 & 0.806 & 0.609 &  \\
 &  & SignaTR6K & 0.618 & 0.715 & 0.449 &  \\
  
 \cmidrule{2-7}
 & \multirow{3}{*}{TEXT+FORM} & DELINE8K & \textbf{0.822} & \textbf{0.83} & \textbf{0.701} &  \\
 &  & SignaTR6K & 0.595 & 0.682 & 0.439 &  \\
 &  & DIBCO & 0.573 & 0.563 & 0.428 &  \\

\cmidrule{2-7}
 & FORM & DELINE8K & \textbf{0.791} & \textbf{0.808} & \textbf{0.655} & 0.7 \\

\cmidrule{1-7}
\multirow{7}{*}{SignaTR6K (test)} 
 & \multirow{3}{*}{HANDWRITING} & SignaTR6K & \textbf{0.985} & \textbf{0.986} & \textbf{0.971} &  \\
 &  & DIBCO & 0.734 & 0.751 & 0.622 &  \\
 &  & DELINE8K & 0.701 & 0.716 & 0.563 &  \\
  
 \cmidrule{2-7}
 & \multirow{3}{*}{TEXT} & SignaTR6K & \textbf{0.949} & \textbf{0.954} & \textbf{0.918} & 0.0 \\
 &  & DELINE8K & 0.44 & 0.436 & 0.334 & 0.4 \\
 &  & DIBCO & 0.251 & 0.254 & 0.19 & 0.0 \\

\cmidrule{2-7}
 & TEXT+FORM & DELINE8K & \textbf{0.62} & \textbf{0.625} & \textbf{0.479} & 2.4 \\

\bottomrule
\end{tabular}
\end{adjustbox}
\end{table}

We did not have access to the models used in~\cite{gholamianHandwrittenPrintedText2023}, though the one we trained appears to have surpassed their performance for both handwriting and printed text, which may be attributable to the additional parameters in our ResNet50 encoder. Unsurprisingly, the models trained on the SignaTR6K training dataset perform the best on the SignaTR6K test dataset, as both sets are drawn from the same distribution of documents. 

The model trained on DELINE8K performs much better on SignaTR6K text than the one trained on DIBCO, though somewhat worse on handwriting overall.

While the model trained on SignaTR6K outperforms DELINE8K on DIBCO overall, DELINE8K far surpasses both DIBCO and SignaTR6K on almost every metric for NAFSS as it was designed to do, despite having to learn the additional form element class.

\subsection{Ablation}

We also perform an ablation comparing the results of DELINE8K with and without the generated backgrounds, with comparative results presented in Table~\ref{tab:ablation_table}. We observe that adding synthetic backgrounds dramatically improves the performance for each class.

\begin{table}[htbp]
\centering
\caption{Performance with and without DALL$\cdot$E synthetic backgrounds}
\label{tab:ablation_table}
\begin{adjustbox}{center, max width=\textwidth}
\begin{tabular}{l@{\hspace{5mm}}l@{\hspace{5mm}}l@{\hspace{5mm}}rr@{\hspace{5mm}}rr}
\toprule
Evaluation Dataset &       Class & Training Dataset &  F-measure &  pF-measure &   IoU &   \\
\midrule
\multirow{8}{*}{NAFSS} & 
\multirow{2}{*}{HANDWRITING} &          DELINE8K &      \textbf{0.876} &       \textbf{0.877} & \textbf{0.785} &    \\
                       &                       &    No Background &      0.808 &       0.804 & 0.687 &    \\
\cmidrule{2-7}
                       & \multirow{2}{*}{TEXT} &          DELINE8K &      \textbf{0.917} &       \textbf{0.914} & \textbf{0.847} &    \\
                       &                       &    No Background &      0.819 &       0.841 & 0.706 &    \\
\cmidrule{2-7}
                       & \multirow{2}{*}{TEXT+FORM} &          DELINE8K &      \textbf{0.822} &       \textbf{0.830} & \textbf{0.701} &    \\
                       &                       &    No Background &      0.750 &       0.779 & 0.604 &    \\
\cmidrule{2-7}
                       & \multirow{2}{*}{FORM} &          DELINE8K &      \textbf{0.791} &       \textbf{0.808} & \textbf{0.655} &    \\
                       &                       &    No Background &      0.683 &       0.715 & 0.532 &    \\
\bottomrule
\end{tabular}
\end{adjustbox}
\end{table}

In summary, the evaluation presented in Table~\ref{tab:MAIN_TABLE} underscores the significant advancement our process brings to the domain of document semantic segmentation. This strong performance is also reflected in the random samples from the NAF dataset visualized in Figure~\ref{fig:NAF_results}. Moreover, we believe the addition of a form elements class to be a valuable contribution that can open the door to unsupervised form classification to distinguish between highly similar forms. 

The use of DALL$\cdot$E-generated backgrounds, as evidenced by the ablation study, not only underscores the effectiveness of synthetic data augmentation in enhancing model performance, but also highlights our innovative approach in overcoming the limitations of traditional synthetic training datasets. We anticipate these techniques playing a significant role moving forward, as image synthesis models improve and our knowledge of how best to apply them grows.

\section{Discussion and Future Work}
The principal advantage of employing synthetic data lies in its ability to be tailored to the specifications of the target dataset. Nonetheless, it should not be considered a comprehensive replacement for real data and there are many areas for improvement. While many of these issues can be partially mitigated using a more targeted synthesized dataset, it may come at a cost of generalizing on other document types.

\begin{figure}[h]
  \centering
  \subfloat[Italicized fonts were mistaken for handwriting in an early synthetic dataset attempt.]{\includegraphics[width=0.75\textwidth]{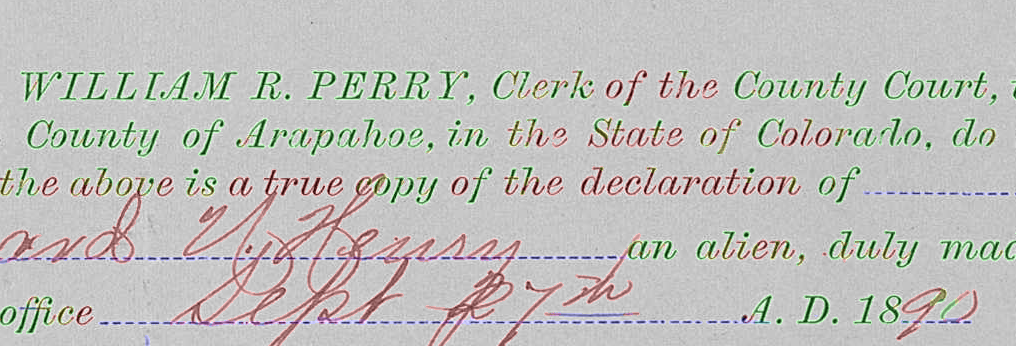}}
    \hspace{5mm}
  \subfloat[Oversampling the most relevant fonts is straightforward with our pipeline. The final DELINE8K dramatically improves this segmentation compared to an earlier version. ]{\includegraphics[width=0.75\textwidth]{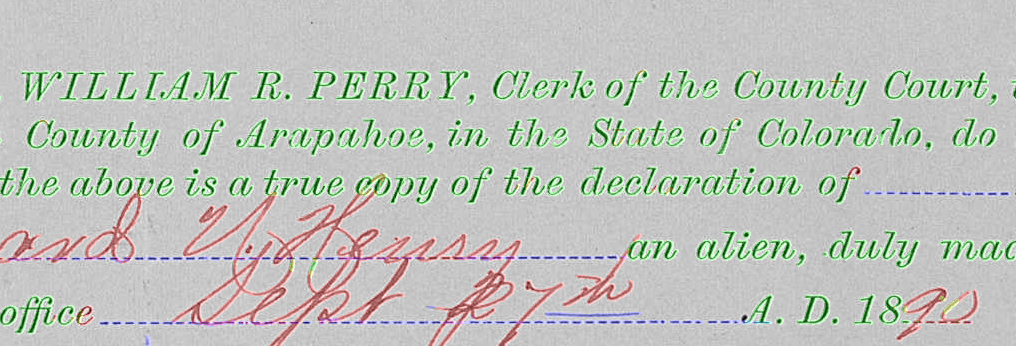}}
  \caption{Synthetic data can be easily adapted to match a target document collection.}
  \label{fig:ITALIC_FAIL}
\end{figure}

\begin{figure}[h]
  \centering
  {\includegraphics[width=.7\textwidth]{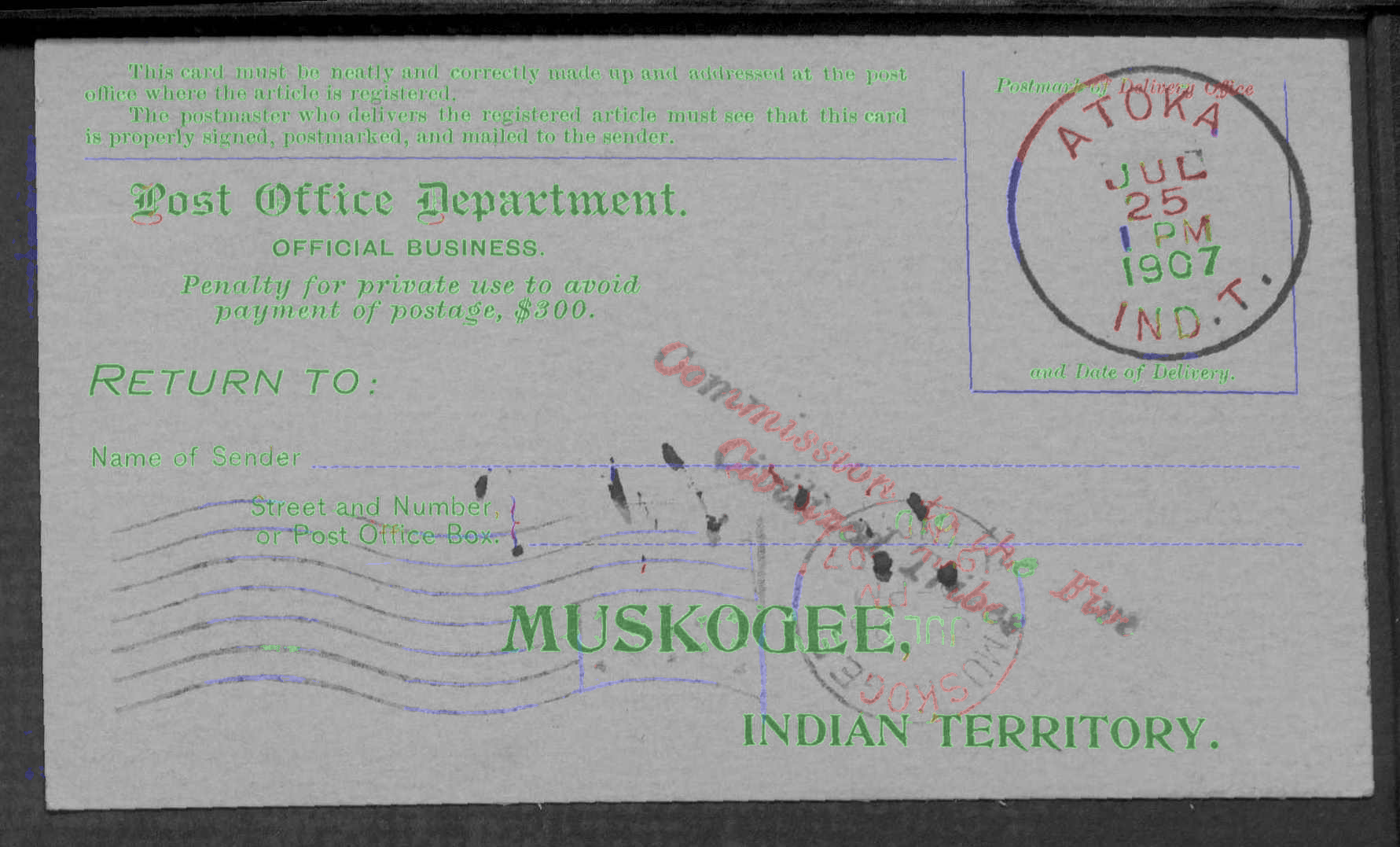}}
  \caption{A postcard presents a particular challenge due to the various fonts used and presences of stamps.}
  \label{fig:MANY_FONTS.png}
\end{figure}

A persistent challenge is distinguishing italic and cursive fonts from handwriting. Figure~\ref{fig:ITALIC_FAIL} demonstrates that using more representative fonts for the target dataset can improve results, though the extent this diminishes handwriting class recognition is unknown. Figure~\ref{fig:MANY_FONTS.png} has at least 10 different font styles and size combinations, which alludes to the limits of creating a synthetic dataset that targets every possible font from a sufficiently diverse target dataset.

\begin{figure}[h]
  \centering
  \subfloat[The model mistakes UNITED STATES OF AMERICA for handwritten text, presumably because the baseline is rounded.]{\includegraphics[width=.42\textwidth]{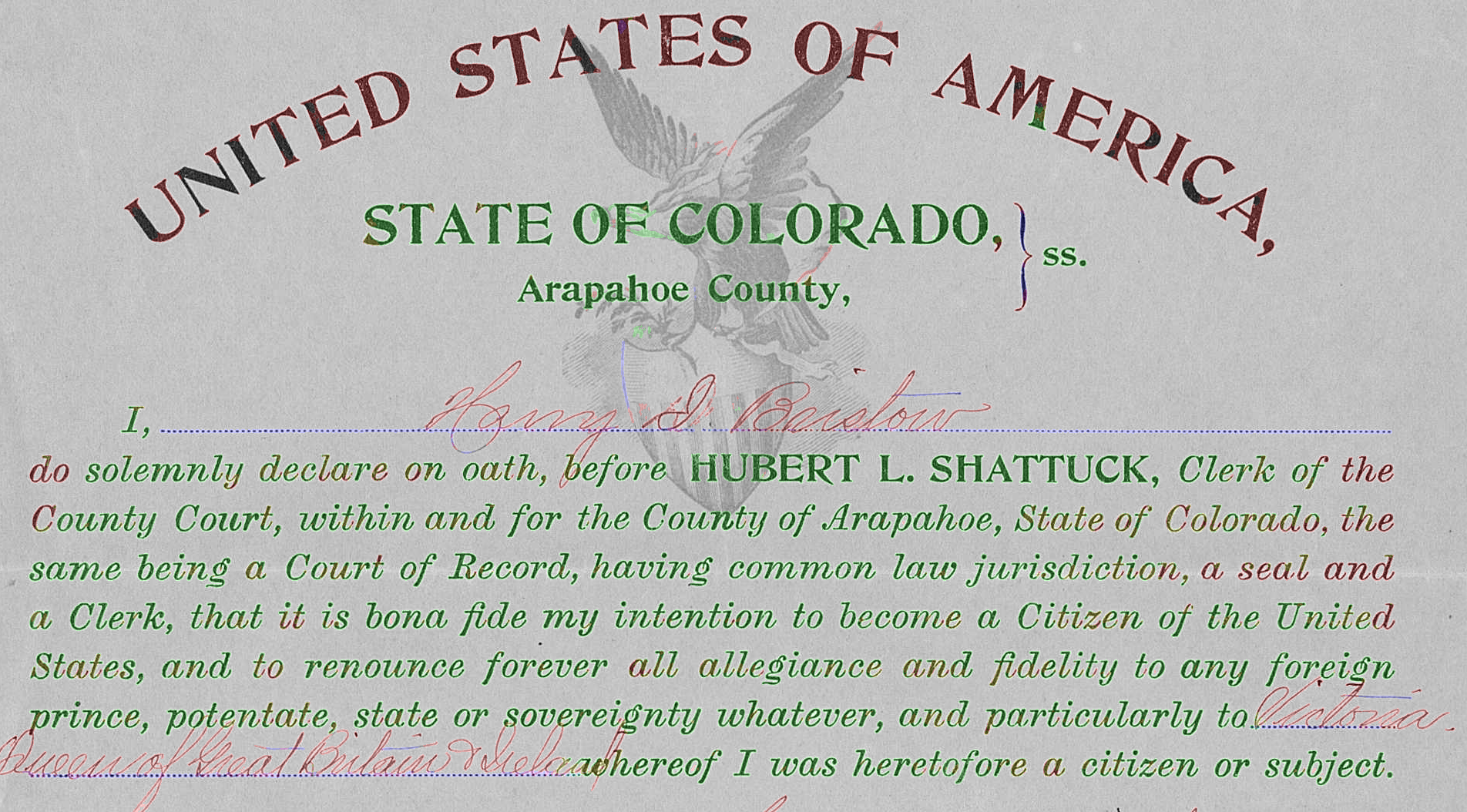}}
  \hspace{5mm}
  \subfloat[A model not trained on vertically-oriented text generalizes poorly.]{\includegraphics[width=0.42\textwidth]{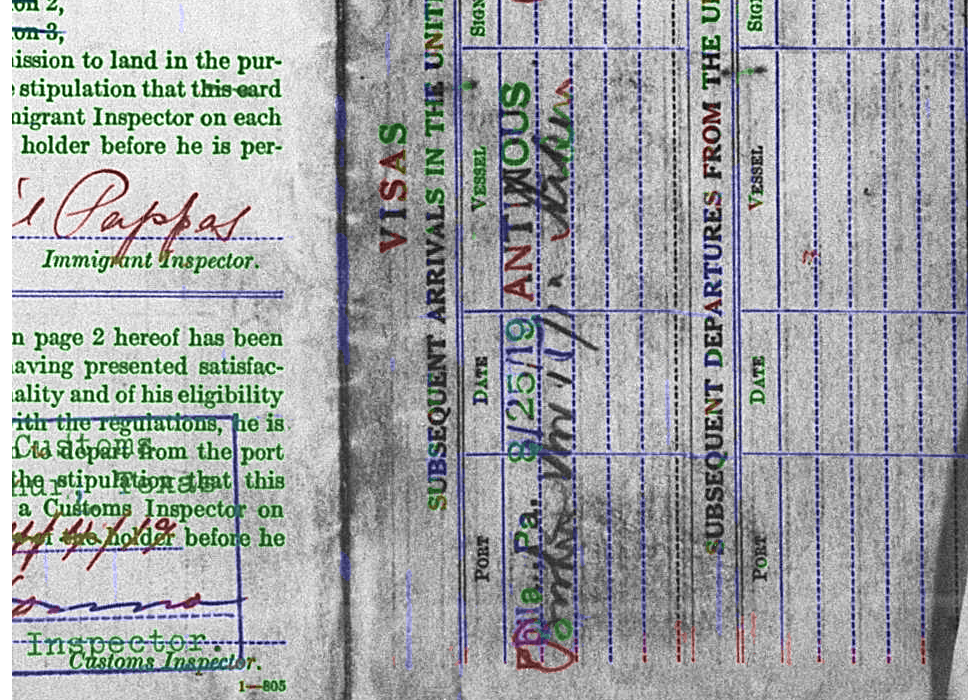}}
  \caption{Nonstandard print orientations should be considered when creating synthetic data.}
  \label{fig:ORIENTATION_FAIL}

\end{figure}

Another potential area for improvement involves training a model that is more robust to nonstandard text alignments, including vertically oriented text or rounded baselines as illustrated in Figure~\ref{fig:ORIENTATION_FAIL}. This was not a focus since these are comparatively rare in the NAF dataset, and training a model to solve these may diminish performance generally.

\begin{figure}[H]
\centering
  {\includegraphics[width=0.5\textwidth]{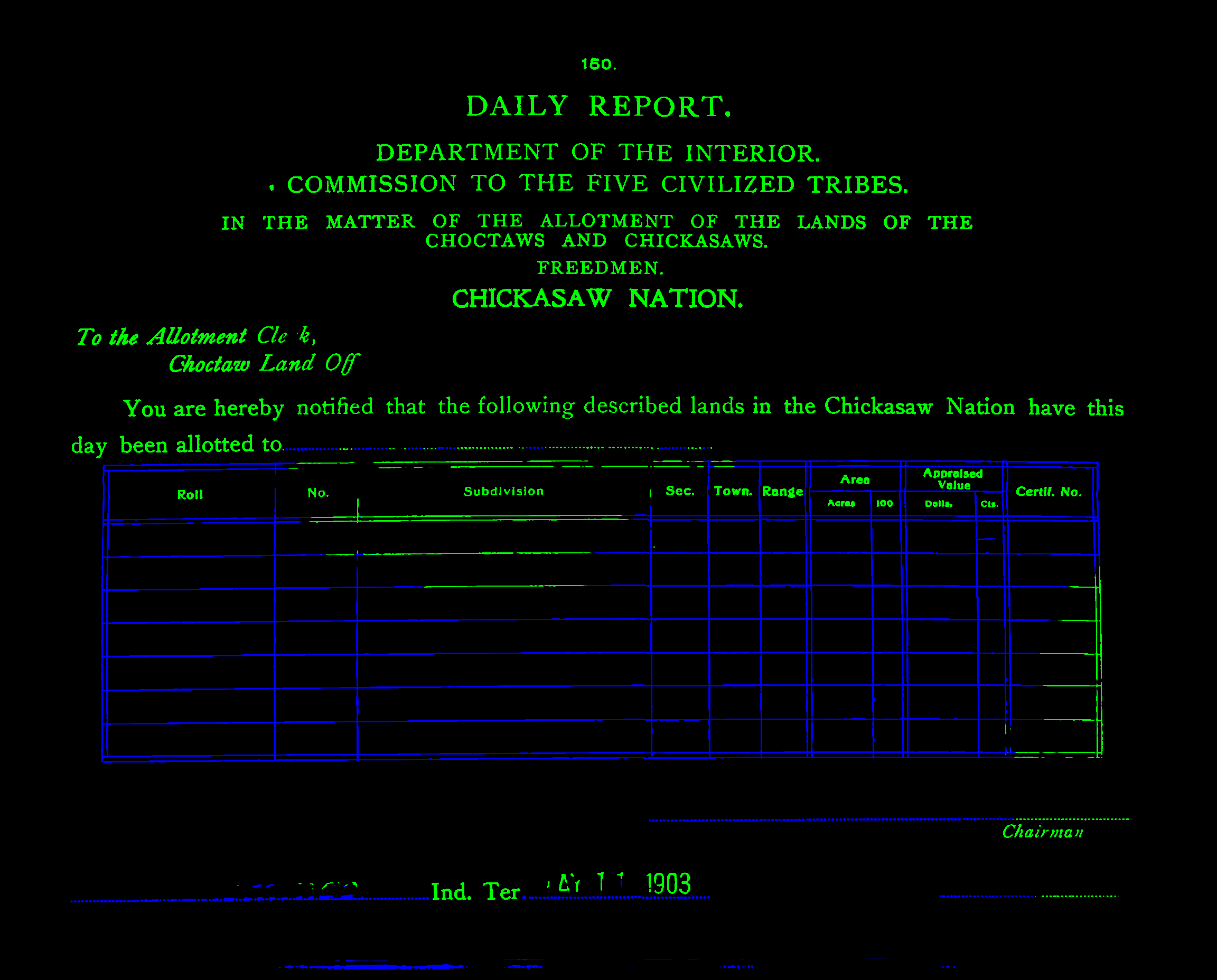}}
  \caption{A model jointly trained on SignaTR6K and DELINE8K yields inconsistent results for the form element class.}
  \label{fig:TEXT_FORM_CONFLATION}

\end{figure}

Jointly training on multiple datasets has the potential to improve the model, but requires more experimentation and must be performed with care. For example, because SignaTR6K does not have labeled form elements, a model jointly trained on SignaTR6K and DELINE8K tends to classify form elements in the same class as text within certain neighborhoods, as seen in Figure~\ref{fig:TEXT_FORM_CONFLATION}. Treating the DELINE8K printed text class as the sum of the model's text and form element class predictions is a potential remedy that can be evaluated.  

Typewritten text and stamps, like handwriting, constitute a dynamic element of a document. Consequently, there may be a preference to distinguish them from pre-printed text. Although stamps frequently emulate the style of printed text, the ink characteristics often bear a closer resemblance to those of handwritten text.

Occasionally, pixels accurately recognized during the binarization process fail to be classified into any specific category. This observation suggests that implementing a two-step procedure—initially binarizing the image followed by the deployment of a classification model—may lead to further enhancements.

\section{Conclusion}
Our investigation into document semantic segmentation demonstrates some of the weaknesses of current synthetic datasets, particularly in terms of class variety and their ability to generalize, as demonstrated by their inferior performance on the NAFSS dataset. Recognizing these challenges, we build a pipeline that let us introduce DELINE8K, a dataset that largely solves the NAFSS dataset. Using DALL$\cdot$E to synthesize diverse document backgrounds for our synthetic dataset dramatically improves performance on NAFSS. Our contribution is not merely DELINE8K, but also the thousands of images, backgrounds, and fonts collected and corresponding code to build targeted synthetic datasets to address future challenges.

\section*{Acknowledgment}

The authors would like to thank the Handwriting Recognition Team at Ancestry.com for providing essential data and support that contributed to the findings of this study. 
\newpage
{
    \small
    \clearpage
    \printbibliography

@online{irsForms,
  author = {{Internal Revenue Service}},
  title = {Forms, Instructions \& Publications},
  year = {2024},
  url = {https://www.irs.gov/forms-instructions-and-publications},
  urldate = {2024-02-14}
}

@online{opmForms,
  author = {{Office of Personnel Management}},
  title = {OPM Forms},
  year = {2024},
  url = {https://www.opm.gov/forms/},
  urldate = {2024-02-14}
}

@online{ssaForms,
  author = {{Social Security Administration}},
  title = {Forms},
  year = {2024},
  url = {https://www.ssa.gov/forms/},
  urldate = {2024-02-14}
}

@online{gsaForms,
  author = {{General Services Administration}},
  title = {GSA Forms},
  year = {2024},
  url = {https://www.gsa.gov/forms},
  urldate = {2024-02-14}
}

@online{freeFonts,
  title = {1001 Free Fonts},
  year = {2024},
  url = {https://www.1001freefonts.com/},
  urldate = {2024-02-14}
}

@misc{ocrodeg,
  author = {{NVlabs}},
  title = {ocrodeg: Document Image Degradation},
  year = {2024},
  publisher = {GitHub},
  journal = {GitHub repository},
  howpublished = {\url{https://github.com/NVlabs/ocrodeg}},
  commit = {The latest commit hash here}
}

@misc{docgen,
  author = {{Taylor Archibald}},
  title = {DocGen},
  year = {2024},
  publisher = {GitHub},
  journal = {GitHub repository},
  howpublished = {\url{https://github.com/Tahlor/docgen}},
  commit = {The latest commit hash here}
}

@inproceedings{bhuniaHandwritingTransformers2021,
	title = {Handwriting Transformers},
	url = {https://openaccess.thecvf.com/content/ICCV2021/html/Bhunia_Handwriting_Transformers_ICCV_2021_paper.html},
	eventtitle = {Proceedings of the {IEEE}/{CVF} International Conference on Computer Vision},
	pages = {1086--1094},
	author = {Bhunia, Ankan Kumar and Khan, Salman and Cholakkal, Hisham and Anwer, Rao Muhammad and Khan, Fahad Shahbaz and Shah, Mubarak},
	urldate = {2024-02-19},
	date = {2021},
	langid = {english},
}

@inproceedings{Wigington2018,
	title = {Data Augmentation for Recognition of Handwritten Words and Lines Using a {CNN}-{LSTM} Network},
	volume = {1},
	isbn = {978-1-5386-3586-5},
	doi = {10.1109/ICDAR.2017.110},
	pages = {639--645},
	booktitle = {Proceedings of the International Conference on Document Analysis and Recognition, {ICDAR}},
	publisher = {{IEEE} Computer Society},
	author = {Wigington, Curtis and Stewart, Seth and Davis, Brian and Barrett, Bill and Price, Brian and Cohen, Scott},
	urldate = {2019-08-28},
	date = {2018-01-25},
	note = {{ISSN}: 15205363},
}

@inproceedings{boilletMultipleDocumentDatasets2021,
  title = {Multiple {{Document Datasets Pre-training Improves Text Line Detection With Deep Neural Networks}}},
  booktitle = {2020 25th {{International Conference}} on {{Pattern Recognition}} ({{ICPR}})},
  author = {Boillet, Mélodie and Kermorvant, Christopher and Paquet, Thierry},
  date = {2021-01-10},
  eprint = {2012.14163},
  eprinttype = {arxiv},
  eprintclass = {cs},
  pages = {2134--2141},
  doi = {10.1109/ICPR48806.2021.9412447},
  url = {http://arxiv.org/abs/2012.14163},
  urldate = {2023-08-16}
}

@article{buslaevAlbumentationsFastFlexible2020,
  title = {Albumentations: {{Fast}} and {{Flexible Image Augmentations}}},
  shorttitle = {Albumentations},
  author = {Buslaev, Alexander and Iglovikov, Vladimir I. and Khvedchenya, Eugene and Parinov, Alex and Druzhinin, Mikhail and Kalinin, Alexandr A.},
  date = {2020-02},
  journaltitle = {Information},
  volume = {11},
  number = {2},
  pages = {125},
  publisher = {{Multidisciplinary Digital Publishing Institute}},
  issn = {2078-2489},
  doi = {10.3390/info11020125},
  url = {https://www.mdpi.com/2078-2489/11/2/125},
  urldate = {2024-02-19},
  issue = {2},
  langid = {english}
}

@inproceedings{cohenEMNISTExtendingMNIST2017,
  title = {{{EMNIST}}: {{Extending MNIST}} to Handwritten Letters},
  shorttitle = {{{EMNIST}}},
  booktitle = {2017 {{International Joint Conference}} on {{Neural Networks}} ({{IJCNN}})},
  author = {Cohen, Gregory and Afshar, Saeed and Tapson, Jonathan and family=Schaik, given=André, prefix=van, useprefix=true},
  date = {2017-05},
  pages = {2921--2926},
  issn = {2161-4407},
  doi = {10.1109/IJCNN.2017.7966217},
  url = {https://ieeexplore.ieee.org/abstract/document/7966217},
  urldate = {2024-02-19},
  eventtitle = {2017 {{International Joint Conference}} on {{Neural Networks}} ({{IJCNN}})}
}

@article{crawfordDatabaseHandwritingSamples2020,
  title = {A Database of Handwriting Samples for Applications in Forensic Statistics},
  author = {Crawford, Amy and Ray, Anyesha and Carriquiry, Alicia},
  date = {2020-02-01},
  journaltitle = {Data in Brief},
  shortjournal = {Data in Brief},
  volume = {28},
  pages = {105059},
  issn = {2352-3409},
  doi = {10.1016/j.dib.2019.105059},
  url = {https://www.sciencedirect.com/science/article/pii/S2352340919314155},
  urldate = {2024-02-19}
}

@online{davisDeepVisualTemplateFree2019,
  title = {Deep {{Visual Template-Free Form Parsing}}},
  author = {Davis, Brian and Morse, Bryan and Cohen, Scott and Price, Brian and Tensmeyer, Chris},
  date = {2019-09-18},
  eprint = {1909.02576},
  eprinttype = {arxiv},
  eprintclass = {cs},
  doi = {10.48550/arXiv.1909.02576},
  url = {http://arxiv.org/abs/1909.02576},
  urldate = {2024-02-12},
  pubstate = {preprint}
}

@online{gholamianHandwrittenPrintedText2023,
  title = {Handwritten and {{Printed Text Segmentation}}: {{A Signature Case Study}}},
  shorttitle = {Handwritten and {{Printed Text Segmentation}}},
  author = {Gholamian, Sina and Vahdat, Ali},
  date = {2023-07-15},
  eprint = {2307.07887},
  eprinttype = {arxiv},
  eprintclass = {cs},
  doi = {10.48550/arXiv.2307.07887},
  url = {http://arxiv.org/abs/2307.07887},
  urldate = {2023-08-18},
  pubstate = {preprint}
}

@online{groleauShabbyPagesReproducibleDocument2023,
  title = {{{ShabbyPages}}: {{A Reproducible Document Denoising}} and {{Binarization Dataset}}},
  shorttitle = {{{ShabbyPages}}},
  author = {Groleau, Alexander and Chee, Kok Wei and Larson, Stefan and Maini, Samay and Boarman, Jonathan},
  date = {2023-03-17},
  eprint = {2303.09339},
  eprinttype = {arxiv},
  eprintclass = {cs},
  doi = {10.48550/arXiv.2303.09339},
  url = {http://arxiv.org/abs/2303.09339},
  urldate = {2023-08-16},
  pubstate = {preprint}
}

@online{heDeepResidualLearning2015,
  title = {Deep {{Residual Learning}} for {{Image Recognition}}},
  author = {He, Kaiming and Zhang, Xiangyu and Ren, Shaoqing and Sun, Jian},
  date = {2015-12-10},
  eprint = {1512.03385},
  eprinttype = {arxiv},
  eprintclass = {cs},
  doi = {10.48550/arXiv.1512.03385},
  url = {http://arxiv.org/abs/1512.03385},
  urldate = {2024-02-17},
  pubstate = {preprint}
}

@article{hullDatabaseHandwrittenText1994,
  title = {A Database for Handwritten Text Recognition Research},
  author = {Hull, J.J.},
  date = {1994-05},
  journaltitle = {IEEE Transactions on Pattern Analysis and Machine Intelligence},
  volume = {16},
  number = {5},
  pages = {550--554},
  issn = {1939-3539},
  doi = {10.1109/34.291440},
  url = {https://ieeexplore.ieee.org/document/291440},
  urldate = {2024-02-19},
  eventtitle = {{{IEEE Transactions}} on {{Pattern Analysis}} and {{Machine Intelligence}}}
}

@article{joHandwrittenTextSegmentation2020,
  title = {Handwritten {{Text Segmentation}} via {{End-to-End Learning}} of {{Convolutional Neural Networks}}},
  author = {Jo, Junho and Koo, Hyung Il and Soh, Jae Woong and Cho, Nam Ik},
  date = {2020-11-01},
  journaltitle = {Multimedia Tools and Applications},
  shortjournal = {Multimed Tools Appl},
  volume = {79},
  number = {43},
  pages = {32137--32150},
  issn = {1573-7721},
  doi = {10.1007/s11042-020-09624-9},
  url = {https://doi.org/10.1007/s11042-020-09624-9},
  urldate = {2023-08-18},
  langid = {english}
}

@inproceedings{kleberCVLDataBaseOffLineDatabase2013,
  title = {{{CVL-DataBase}}: {{An Off-Line Database}} for {{Writer Retrieval}}, {{Writer Identification}} and {{Word Spotting}}},
  shorttitle = {{{CVL-DataBase}}},
  booktitle = {2013 12th {{International Conference}} on {{Document Analysis}} and {{Recognition}}},
  author = {Kleber, Florian and Fiel, Stefan and Diem, Markus and Sablatnig, Robert},
  date = {2013-08},
  pages = {560--564},
  publisher = {{IEEE}},
  location = {{Washington, DC, USA}},
  doi = {10.1109/ICDAR.2013.117},
  url = {http://ieeexplore.ieee.org/document/6628682/},
  urldate = {2024-02-19},
  eventtitle = {2013 12th {{International Conference}} on {{Document Analysis}} and {{Recognition}} ({{ICDAR}})},
  isbn = {978-0-7695-4999-6},
  langid = {english}
}

@report{martiIAMdatabaseEnglishSentence2002,
  title = {The {{IAM-database}}: An {{English}} Sentence Database for Offline Handwriting Recognition},
  author = {Marti, U.-V and Bunke, H},
  date = {2002},
  volume = {5},
  pages = {39--46},
  url = {http://www.tummy.com/xvscan/},
  urldate = {2020-07-23}
}

@online{ObjectiveEvaluationMethodologyBinarization,
  title = {An {{Objective Evaluation Methodology}} for {{Document Image Binarization Techniques}} | {{IEEE Conference Publication}} | {{IEEE Xplore}}},
  url = {https://ieeexplore.ieee.org/abstract/document/4669964},
  urldate = {2024-02-19}
}

@article{perezPoissonImageEditing2003,
  title = {Poisson Image Editing},
  author = {Pérez, Patrick and Gangnet, Michel and Blake, Andrew},
  date = {2003-07-01},
  journaltitle = {ACM Transactions on Graphics},
  shortjournal = {ACM Trans. Graph.},
  volume = {22},
  number = {3},
  pages = {313--318},
  issn = {0730-0301},
  doi = {10.1145/882262.882269},
  url = {https://dl.acm.org/doi/10.1145/882262.882269},
  urldate = {2024-02-19}
}

@inproceedings{pratikakisICDAR2017CompetitionDocument2017,
  title = {{{ICDAR2017 Competition}} on {{Document Image Binarization}} ({{DIBCO}} 2017)},
  booktitle = {2017 14th {{IAPR International Conference}} on {{Document Analysis}} and {{Recognition}} ({{ICDAR}})},
  author = {Pratikakis, Ioannis and Zagoris, Konstantinos and Barlas, George and Gatos, Basilis},
  date = {2017-11},
  volume = {01},
  pages = {1395--1403},
  issn = {2379-2140},
  doi = {10.1109/ICDAR.2017.228},
  eventtitle = {2017 14th {{IAPR International Conference}} on {{Document Analysis}} and {{Recognition}} ({{ICDAR}})}
}

@inproceedings{pratikakisICFHR2012Competition2012,
  title = {{{ICFHR}} 2012 {{Competition}} on {{Handwritten Document Image Binarization}} ({{H-DIBCO}} 2012)},
  booktitle = {2012 {{International Conference}} on {{Frontiers}} in {{Handwriting Recognition}}},
  author = {Pratikakis, Ioannis and Gatos, Basilis and Ntirogiannis, Konstantinos},
  date = {2012-09},
  pages = {817--822},
  doi = {10.1109/ICFHR.2012.216},
  url = {https://ieeexplore.ieee.org/abstract/document/6424498},
  urldate = {2024-02-19},
  eventtitle = {2012 {{International Conference}} on {{Frontiers}} in {{Handwriting Recognition}}}
}

@report{rameshZeroShotTexttoImageGeneration,
  title = {Zero-{{Shot Text-to-Image Generation}}},
  author = {Ramesh, Aditya and Pavlov, Mikhail and Goh, Gabriel and Gray, Scott and Voss, Chelsea and Radford, Alec and Chen, Mark and Sutskever, Ilya},
  eprint = {2102.12092v1},
  eprinttype = {arxiv},
  url = {https://github.com/openai/DALL-E},
  urldate = {2021-02-26}
}

@online{ronnebergerUNetConvolutionalNetworks2015,
  title = {U-{{Net}}: {{Convolutional Networks}} for {{Biomedical Image Segmentation}}},
  shorttitle = {U-{{Net}}},
  author = {Ronneberger, Olaf and Fischer, Philipp and Brox, Thomas},
  date = {2015-05-18},
  eprint = {1505.04597},
  eprinttype = {arxiv},
  eprintclass = {cs},
  doi = {10.48550/arXiv.1505.04597},
  url = {http://arxiv.org/abs/1505.04597},
  urldate = {2023-08-17},
  pubstate = {preprint}
}

@inproceedings{sadekarLSHDIBLargeScale2022,
  title = {{{LS-HDIB}}: {{A Large Scale Handwritten Document Image Binarization Dataset}}},
  shorttitle = {{{LS-HDIB}}},
  booktitle = {2022 26th {{International Conference}} on {{Pattern Recognition}} ({{ICPR}})},
  author = {Sadekar, Kaustubh and Tiwari, Ashish and Singh, Prajwal and Raman, Shanmuganathan},
  date = {2022-08},
  pages = {1678--1684},
  issn = {2831-7475},
  doi = {10.1109/ICPR56361.2022.9956447},
  eventtitle = {2022 26th {{International Conference}} on {{Pattern Recognition}} ({{ICPR}})}
}

@inproceedings{stewartDocumentImagePage2017,
  title = {Document {{Image Page Segmentation}} and {{Character Recognition}} as {{Semantic Segmentation}}},
  booktitle = {Proceedings of the 4th {{International Workshop}} on {{Historical Document Imaging}} and {{Processing}}},
  author = {Stewart, Seth and Barrett, Bill},
  date = {2017-11-10},
  series = {{{HIP}} '17},
  pages = {101--106},
  publisher = {{Association for Computing Machinery}},
  location = {{New York, NY, USA}},
  doi = {10.1145/3151509.3151518},
  url = {https://dl.acm.org/doi/10.1145/3151509.3151518},
  urldate = {2024-01-31},
  isbn = {978-1-4503-5390-8}
}

@inproceedings{tensmeyerDocumentImageBinarization2017,
  title = {Document {{Image Binarization}} with {{Fully Convolutional Neural Networks}}},
  booktitle = {2017 14th {{IAPR International Conference}} on {{Document Analysis}} and {{Recognition}} ({{ICDAR}})},
  author = {Tensmeyer, Chris and Martinez, Tony},
  date = {2017-11},
  volume = {01},
  pages = {99--104},
  issn = {2379-2140},
  doi = {10.1109/ICDAR.2017.25},
  eventtitle = {2017 14th {{IAPR International Conference}} on {{Document Analysis}} and {{Recognition}} ({{ICDAR}})}
}

@software{theaugraphyprojectAugraphyAugmentationPipeline2023,
  title = {Augraphy: An Augmentation Pipeline for Rendering Synthetic Paper Printing, Faxing, Scanning and Copy Machine Processes},
  shorttitle = {Augraphy},
  author = {{The Augraphy Project}},
  date = {2023-08-21T09:07:25Z},
  origdate = {2021-05-31T16:27:41Z},
  url = {https://github.com/sparkfish/augraphy},
  urldate = {2023-08-22},
  version = {8.2}
}

@article{vafaieHandwrittenPrintedText2022,
  title = {Handwritten and {{Printed Text Identification}} in {{Historical Archival Documents}}},
  author = {Vafaie, Mahsa and Bruns, Oleksandra and Pilz, Nastasja and Waitelonis, Jörg and Sack, Harald and Bruns, Oleksandra and Pilz, Nastasja and Waitelonis, Jörg and Sack, Harald},
  date = {2022-06-07},
  journaltitle = {Archiving Conference},
  volume = {19},
  pages = {15--20},
  publisher = {{Society for Imaging Science and Technology}},
  issn = {2161-8798},
  doi = {10.2352/issn.2168-3204.2022.19.1.4},
  url = {https://library.imaging.org/archiving/articles/19/1/4},
  urldate = {2023-08-18},
  langid = {english}
}
    \newpage
}
\newpage
\section*{Supplemental Materials}  
\appendix
\renewcommand{\appendixname}{Supplemental Section}

\section{Data augmentation}\label{augmentation}
In our image preprocessing pipeline, we employed a diverse set of augmentation strategies to enhance model robustness against various image qualities and artifacts during training.

\subsection{Albumentations}
We split augmentations into 2 phases: augmentations that apply only to the input image, and those that apply to both the image and the label.

\begin{itemize}
\item \textbf{Phase 1:}
    \begin{itemize}
    \item \textbf{Random Brightness Contrast:} Adjusts brightness and contrast with limits set to 0.1, applied with a probability of 0.2.
    \item \textbf{Blur:} Applies blur with a limit of 3, applied with a probability of 0.1.
    \item \textbf{Gaussian Noise:} Introduces Gaussian noise with a probability of 0.3.
    \end{itemize}
\item \textbf{Phase 2:}
    \begin{itemize}
        \item \textbf{Rotate:} Rotates images within $\pm10$ degrees and $\pm90$ degrees, with different probabilities (0.2 for $\pm10$, 0.01 for $\pm90$, and $-90$ degrees rotations).
        \item \textbf{Random Scale:} Scales images within a limit of -0.5 to 0.2, with a probability of 0.7.
        \item \textbf{Random Crop:} Crops images to $448\times448$ pixels.
        \item \textbf{Shift Scale Rotate:} Applies slight rotation, scaling, and translation with a limit of 10 degrees for rotation, applied with a probability of 0.1.
    \end{itemize}
\end{itemize}

\subsection{Augraphy Augmentations}

Our dataset augmentation included simulating various document and image quality challenges such as BleedThrough, ShadowCast, NoiseTexturize, BrightnessTexturize, DirtyDrum, LowInkPeriodicLines, InkBleed, ImageRotator, MeshGridWarp, SaltAndPepperNoiseAdder, and others, each with specified probabilities to mimic real-world imperfections in scanned or photographed documents.

\subsection{Other Augmentations}
In the creation of our dataset, we used multiscale noise from~\cite{ocrodeg} to generate noisy backgrounds. During training, we sometimes call RandomGridWarp from~\cite{Wigington2018}.

\section{DALL$\cdot$E Generation}

Hundreds of prompts were tested to produce consistently usable results. Some of those used are included below.

\subsection*{Prompts that typically yielded a document in the foreground with significant background elements}

\begin{itemize}
    \item Full frame aerial view of 2 blank pages of old open book with wrinkles; book is on top of a noisy background
    \item Microfilm blank old page with ink marks
    \item Old blank weathered document
    \item Old microfilm blank document
    \item 2 blank pages of old book, full frame
    \item Aerial view of 2 blank pages of old open book with ink marks and wrinkles, full frame
    \item Aerial view of 2 blank pages of old open book, full frame
    \item Blank discolored paper with creases; it is stapled along the top, bottom, and sides, also has paperclips
    \item Blank old document
    \item Old paper with many ink marks, crinkles, wrinkles, and imperfections and variable lighting
    \item Paper with many ink marks, crinkles, wrinkles, and imperfections and variable lighting
    \item White paper with many ink marks, crinkles, wrinkles, and imperfections and variable lighting
\end{itemize}

\subsection*{Prompts that produced document images with minimal backgrounds}

\begin{itemize}
    \item Aged blank letter with imprints of text from the reverse side, as if some vestiges of the ink permeated through the paper, but impossible to read; full frame
    \item Aged blank letter with subtle imprints of text from its other side, as if the ink permeated through the paper; full frame
    \item Blank paper with some random highlighter and marker marks, full frame
    \item Blank paper with mold damage, full frame
    \item Old blank paper with some random highlighter and marker marks, full frame
    \item Old blank paper with water or coffee stains, full frame
    \item Old blank paper, some wrinkles and imperfections, variable lighting
    \item Old paper with ink marks, crinkles, wrinkles, and imperfections and variable lighting
\end{itemize}

The images often contained artifacts. For instance, blank documents with coffee stains often also included a drawing of a coffee mug on the page if not a rendering a realistic coffee mug resting on top of the document.




\begin{figure}[h]
  \centering
  {\includegraphics[width=0.7\textwidth]{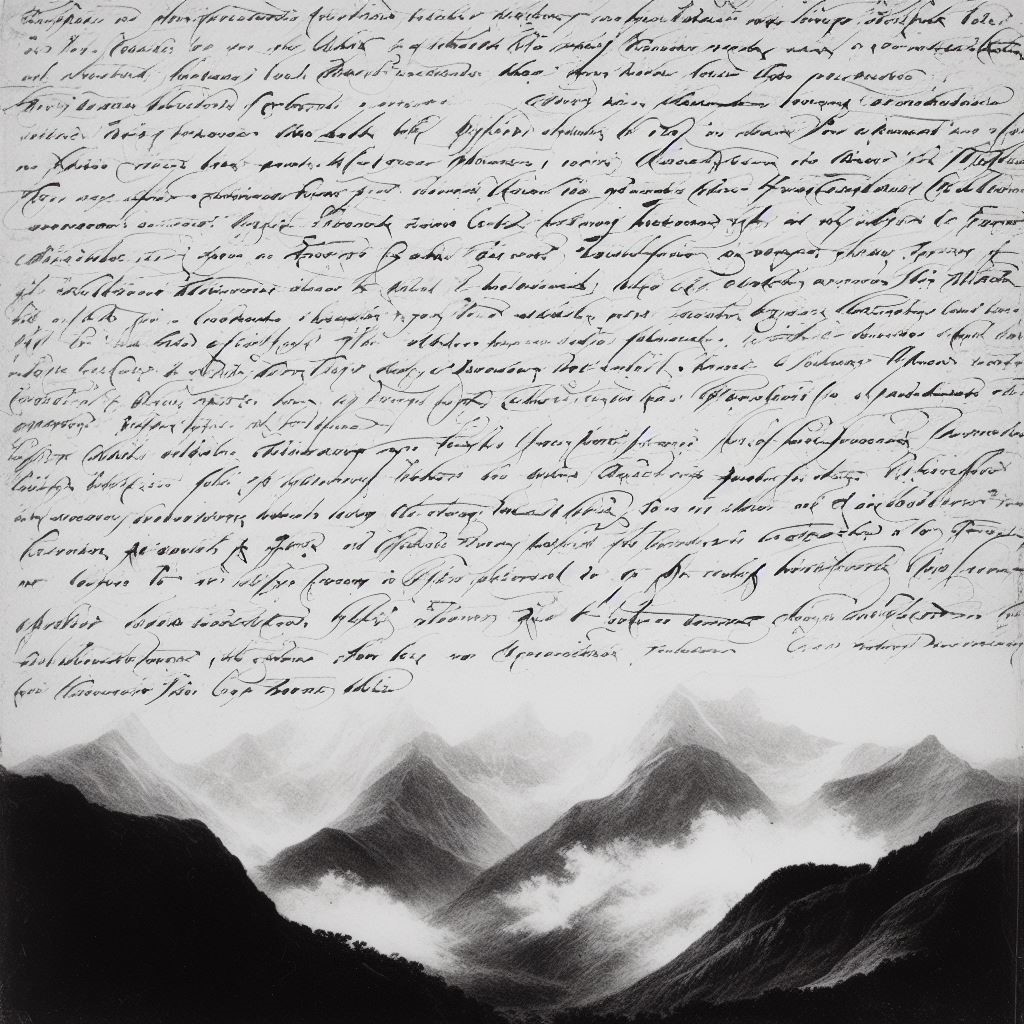}}
  \caption{A handwritten document generated by DALL$\cdot$E.}
  \label{fig:mountains}

\end{figure}

Using DALL$\cdot$E to generate documents mimicking ink bleedthrough did not seem to improve results. We found it difficult to get DALL$\cdot$E to generate clean handwriting samples. While these were diverse in terms of language, they lacked diversity in many other respects, including spacing, size, and ink quality. Often, handwritten documents contained sketches of mountains, as in Figure~\ref{fig:mountains}. 

Prompts included:

\begin{itemize}
    \item a black and white handwritten manuscript; the first 5 lines of a handwritten essay; a scanned document; pristine pure white paper background; image is square, centered, high contrast, no flourishes
    \item a black and white handwritten manuscript; the first 5 lines of a handwritten essay; very large words; a scanned document; pristine pure white paper background; image is square, centered, high contrast
    \item a black and white handwritten manuscript with only English 19th century text; a scanned document; pristine pure white paper background; image is square, centered, high contrast, no flourishes, no pictures
    \item a black and white handwritten manuscript with only English text; a scanned document; pristine pure white paper background; image is square, centered, high contrast, no flourishes, no pictures, no drawing
    \item a black and white handwritten manuscript with only text; a scanned document; pristine pure white paper background; image is square, centered, high contrast, no flourishes, no pictures, no drawings
    \item a black and white handwritten manuscript; a scanned document; pristine pure white paper background; image is square, centered, high contrast, no flourishes, no pictures, no drawings, no shadows
\end{itemize}

\end{document}